\documentclass{article}
\usepackage{amsmath,epsfig}
\usepackage[preprint]{spconfa4}
\usepackage{xcolor}
\usepackage{cite}
\usepackage{amssymb}
\usepackage{subcaption}
\usepackage{booktabs} 
\usepackage{multirow}
\usepackage{makecell}

\let\OLDthebibliography\thebibliography
\renewcommand\thebibliography[1]{
  \OLDthebibliography{#1}
  \setlength{\parskip}{0pt}
  \setlength{\itemsep}{0pt plus 0.3ex}
}

\begin{document}\sloppy

\def\x{{\mathbf x}}
\def\L{{\cal L}}

\title{Efficient CNN Architecture Design Guided by Visualization}
%
\name{Liangqi Zhang, Haibo Shen, Yihao Luo, Xiang Cao, Leixilan Pan, Tianjiang Wang$^{\ast}$\thanks{* Corresponding Author}, Qi Feng}
\address{School of Computer Science and Technology, Huazhong University of Science and Technology, China \\ \{zhangliangqi, shenhaibo, luoyihao, caoxiang112, d202081082, tjwang, fengqi\}@hust.edu.cn}

\maketitle

\begin{abstract}
	Modern efficient Convolutional Neural Networks(CNNs) always use Depthwise Separable Convolutions(DSCs) and Neural Architecture Search(NAS) to reduce the number of parameters and the computational complexity.
	But some inherent characteristics of networks are overlooked.
	Inspired by visualizing feature maps and N$\times$N(N$>$1) convolution kernels,
	several guidelines are introduced in this paper to further improve parameter efficiency and inference speed.
	Based on these guidelines, our parameter-efficient CNN architecture, called \textit{VGNetG},
	achieves better accuracy and lower latency than previous networks with about 30\%$\thicksim$50\% parameters reduction.
	Our VGNetG-1.0MP achieves 67.7\% top-1 accuracy with 0.99M parameters and
	69.2\% top-1 accuracy with 1.14M parameters on ImageNet classification dataset.

	Furthermore, we demonstrate that edge detectors can replace learnable depthwise convolution layers to mix features by replacing the N$\times$N kernels with fixed edge detection kernels.
	And our VGNetF-1.5MP archives 64.4\%(-3.2\%) top-1 accuracy and 66.2\%(-1.4\%) top-1 accuracy with additional Gaussian kernels.

\end{abstract}
\begin{keywords}
	Visualization, Efficient, Edge detection, Gaussian blur
\end{keywords}
\section{Introduction}
\label{sec:intro}

Recently, Convolutional Neural Networks(CNNs) have made great progress in computer vision.
Since AlexNet\cite{Krizhevsky2012}, CNN-based methods focus on designing wider or deeper network architectures for the accuracy gains,
including VGGNets\cite{Simonyan2015}, ResNets\cite{He2016}, and DenseNets\cite{Huang2017}.
However, the computational and storage capacity is always limited.
The most prominent approaches to reduce the parameters and computational complexity are based on
\textit{Depthwise Separable Convolutions}\cite{Sifre2014} and \textit{Neural Architecture Search},
such as MobileNets\cite{Howard2017,Sandler2018}, ShuffleNets\cite{Zhang2018,Ma2018}, and EfficientNets\cite{Tan2019a}.
Although these approaches have been greatly successful, they also overlook many inherent characteristics of convolutional neural networks.

Visualization is a powerful tool to study neural networks. \textit{Visualization of features} in a fully trained model
\cite{Erhan2009,Zeiler2014} makes us to see the process of extracting features.
\textit{Feature Visualization by Optimization} \cite{Zeiler2011,Yosinski2015} explains what a
network is looking for. \textit{Attribution} \cite{Simonyan2014,Springenberg2015,Selvaraju2020}
explains what part of an example is responsible for the network activating in a particular way.
Visualization can reveal many inherent characteristics of neural networks.

\begin{figure}[t]
	\centering
	\begin{minipage}[b]{.19\linewidth}
		\centering
		\includegraphics[width=.96\linewidth]{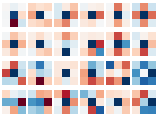}
		\includegraphics[width=.96\linewidth]{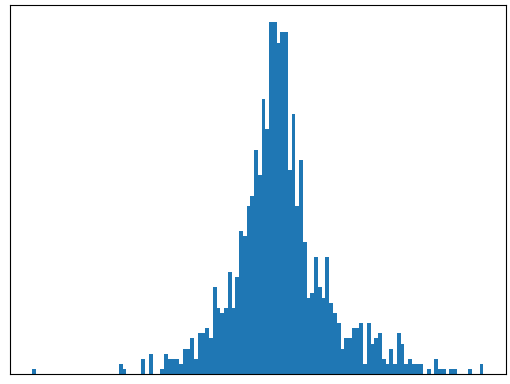}
		\captionsetup{justification=centering}
		\subcaption{\#3/s = 1\\ks=1$\times$3$\times$3}\label{fig:examples-1}
	\end{minipage}
	\begin{minipage}[b]{.19\linewidth}
		\centering
		\includegraphics[width=.96\linewidth]{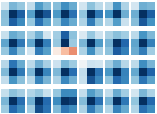}
		\includegraphics[width=.96\linewidth]{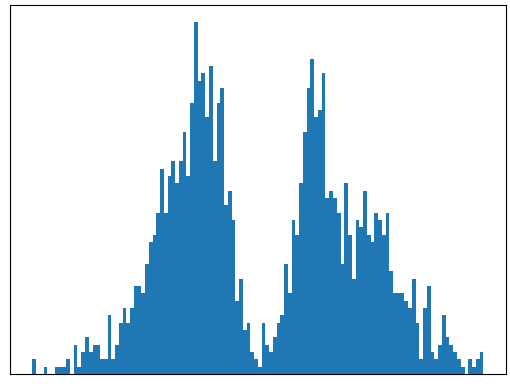}
		\captionsetup{justification=centering}
		\subcaption{\#7/s = 2\\ks=1$\times$3$\times$3}\label{fig:examples-2}
	\end{minipage}
	\begin{minipage}[b]{.19\linewidth}
		\centering
		\includegraphics[width=.96\linewidth]{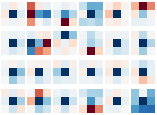}
		\includegraphics[width=.96\linewidth]{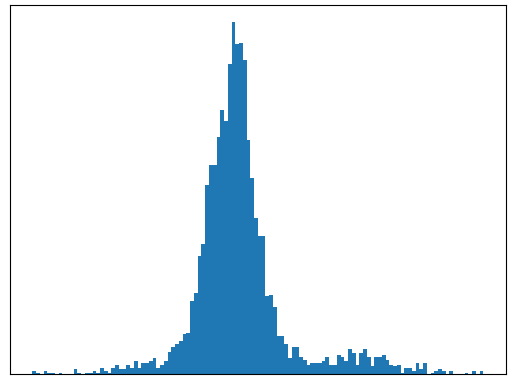}
		\captionsetup{justification=centering}
		\subcaption{\#11/s = 1\\ks=1$\times$3$\times$3}\label{fig:examples-3}
	\end{minipage}
	\begin{minipage}[b]{.19\linewidth}
		\centering
		\includegraphics[width=.96\linewidth]{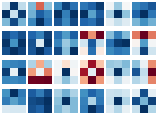}
		\includegraphics[width=.96\linewidth]{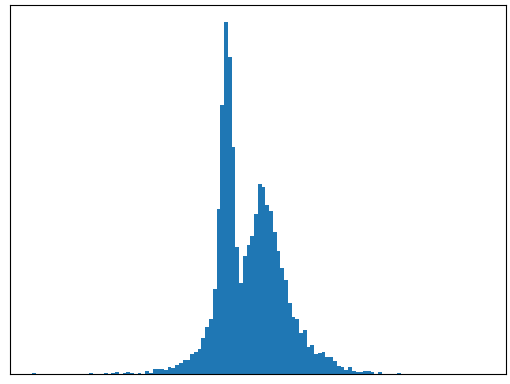}
		\captionsetup{justification=centering}
		\subcaption{\#16/s = 1\\ks=1$\times$3$\times$3}\label{fig:examples-4}
	\end{minipage}
	\begin{minipage}[b]{.19\linewidth}
		\centering
		\includegraphics[width=.96\linewidth]{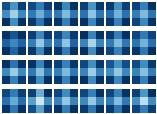}
		\includegraphics[width=.96\linewidth]{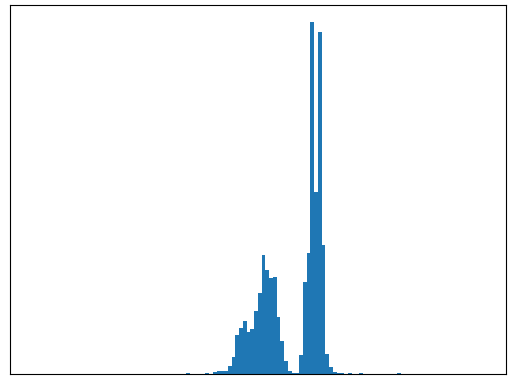}
		\captionsetup{justification=centering}
		\subcaption{last/s = 1\\ks=1$\times$3$\times$3}\label{fig:examples-5}
	\end{minipage}
	\setlength{\abovecaptionskip}{0.1cm}
	\captionsetup{font={small}}
	\caption{
		\textbf{Kerenls and their distributions of MobileNetV2}.
		The kernels at different stages in the network show distinctly different patterns,
		like edge detection filter kernels, blur kernels or identity kernels.
		Here, in \#N, \textit{N} denotes layer number, \textit{s} denotes stride and \textit{ks} denotes kernel size.
	}
	\label{fig:kernels_stages}
\end{figure}

In this paper, we study the characteristics of networks by visualizing the N$\times$N kernels, the distribution of kernels, and feature maps.
As shown in Figure \ref{fig:kernels_stages}, the N$\times$N convolution kernels show distinctly different patterns and distributions at different stages of MobileNetV2\cite{Sandler2018}.

Our VGNets guided by these visualizations achieve better accuracy and lower latency than previous models with about 30\%$\thicksim$50\% parameters reduction.
Specifically, Our VGNetG-1.0MP archives 67.7\% top-1 accuracy with 0.997M parameters without strong regularization methods.
Furthermore, we demonstrate that edge detectors can replace learnable depthwise convolutions for mixing features between different spatial locations.

\begin{figure*}[ht]
	\centering
	\small
	\begin{minipage}{0.97\textwidth}
		\includegraphics[width=\linewidth]{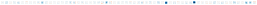}
		\includegraphics[width=\linewidth]{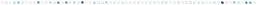}
		\includegraphics[width=\linewidth]{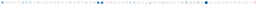}
		\subcaption{ResNet-RS 50: \#18/64$\times$3$\times$3/s=2}\label{fig:resnet_rs_50}
	\end{minipage}

	\begin{minipage}{0.97\textwidth}
		\includegraphics[width=\linewidth]{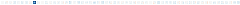}
		\includegraphics[width=\linewidth]{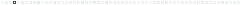}
		\includegraphics[width=\linewidth]{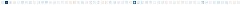}
		\subcaption{RegNetX-8GF: \#24/120$\times$3$\times$3(partial)/s=2}\label{fig:regnetx8gf_downsampling}
	\end{minipage}

	\begin{minipage}{0.33\textwidth}
		\centering
		\includegraphics[width=.71\linewidth]{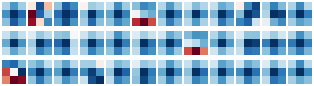}
		\includegraphics[width=.26\linewidth]{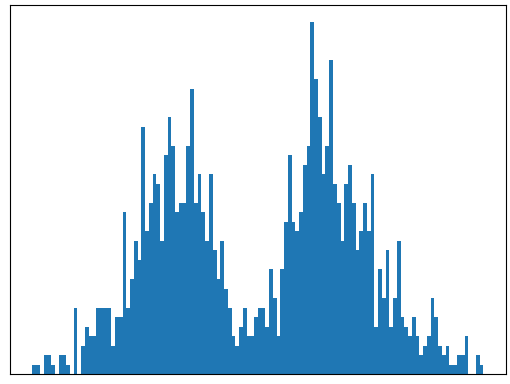}
		\subcaption{MobileNetV2: \#4/1$\times$3$\times$3}\label{fig:mobilenetv2_c4s2}
	\end{minipage}
	\begin{minipage}{0.33\textwidth}
		\centering
		\includegraphics[width=.71\linewidth]{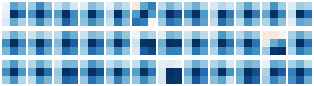}
		\includegraphics[width=.26\linewidth]{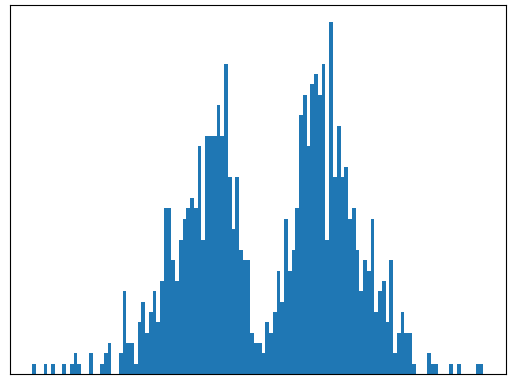}
		\subcaption{ShuffleNetV2: \#6/1$\times$3$\times$3}\label{fig:shufflenet_v2_x1_0_c6s2}
	\end{minipage}
	\begin{minipage}{0.33\textwidth}
		\centering
		\includegraphics[width=.71\linewidth]{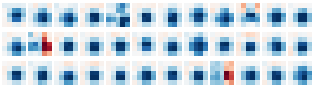}
		\includegraphics[width=.26\linewidth]{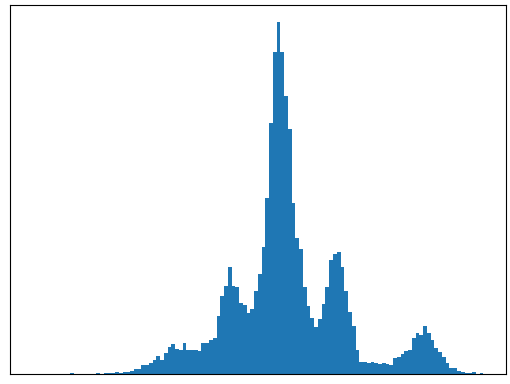}
		\subcaption{EfficientNet-B0: \#12/1$\times$5$\times$5}\label{fig:efficientnet_b7_c39s2}
	\end{minipage}
	\setlength{\abovecaptionskip}{0.1cm}
	\setlength{\belowcaptionskip}{-0.1cm}
	\captionsetup{font={small}}
	\caption{\textbf{The kernels of downsampling layers are similar to low-pass filter kernels.}
		(a) Standard convolution kernels and (b) Group convolution kernels:
		have one or more salient N$\times$N kernels like blur kernels in the whole M$\times$N$\times$N kernel.
		(c)(d)(e) Depthwise convolution kernels: like blur kernels, especially Gaussian kernels.
	}
	\label{fig:gaussian}
\end{figure*}

\begin{figure*}[ht]
	\centering
	\small
	\begin{minipage}{.24\linewidth}
		\centering
		\includegraphics[width=.97\linewidth]{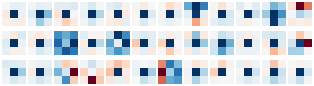}
		\subcaption{MobileNetV2: \#11/$1 \times 3 \times 3$}
	\end{minipage}
	\begin{minipage}{.24\linewidth}
		\centering
		\includegraphics[width=.97\linewidth]{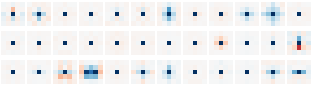}
		\subcaption{MobiletNetV3-S: \#7/$1 \times 5 \times 5$}
	\end{minipage}
	\begin{minipage}{.24\linewidth}
		\centering
		\includegraphics[width=.97\linewidth]{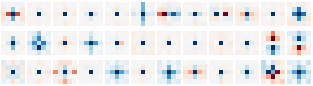}
		\subcaption{EfficientNet-B0: \#9/$1 \times 5 \times 5$}
	\end{minipage}
	\begin{minipage}{.24\linewidth}
		\centering
		\includegraphics[width=.97\linewidth]{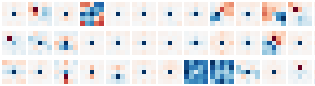}
		\subcaption{EfficientNet-B7: \#29/$1 \times 5 \times 5$}
	\end{minipage}
	\setlength{\abovecaptionskip}{0.1cm}

	\captionsetup{font={small}}
	\caption{
		\textbf{Kernels from the middle layers of the network}.
		It can be seen that lots of convolution kernels are similar to \textit{the identity kernel}.
		As a result, we replace some convolution operations with identity mapping to reduce the parameters and the computation complexity.
	}
	\label{fig:identity}
\end{figure*}

\begin{figure*}[ht]
	\centering
	\small
	\begin{minipage}[b]{.33\linewidth}
		\centering
		\includegraphics[width=.71\linewidth]{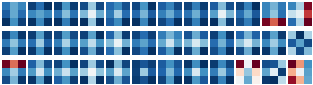}
		\includegraphics[width=.26\linewidth]{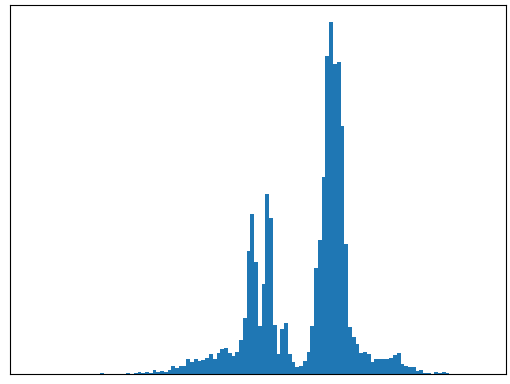}
		\subcaption{MobileNetV2: ks=1$\times$3$\times$3}
	\end{minipage}
	\begin{minipage}[b]{.33\linewidth}
		\centering
		\includegraphics[width=.71\linewidth]{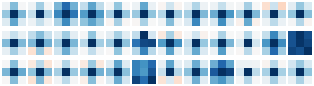}
		\includegraphics[width=.26\linewidth]{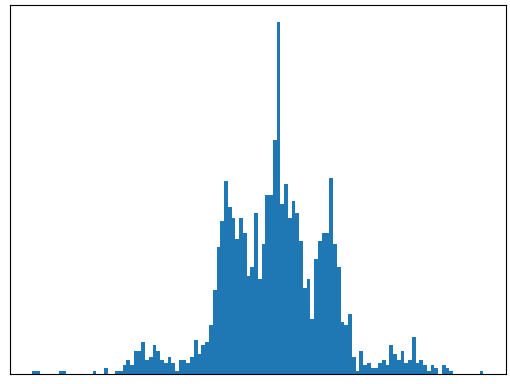}
		\subcaption{ShuffleNetV2: ks=1$\times$3$\times$3}
	\end{minipage}
	\begin{minipage}[b]{0.33\textwidth}
		\begin{minipage}[b]{.71\linewidth}
			\includegraphics[width=\linewidth]{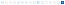}
			\includegraphics[width=\linewidth]{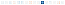}
			\includegraphics[width=\linewidth]{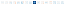}
		\end{minipage}
		\includegraphics[width=.26\linewidth]{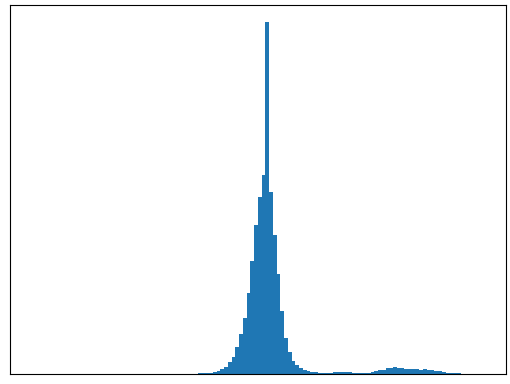}
		\subcaption{RegNetX-800MF: ks=16$\times$3$\times$3}
	\end{minipage}
	\setlength{\abovecaptionskip}{-0.1cm}
	\setlength{\belowcaptionskip}{-0.15cm}
	\captionsetup{font={small}}
	\caption{The M$\times$N$\times$N kernels of last convolution layers show similar phenomenon.}
	\label{fig:last}
\end{figure*}

\begin{figure}[ht]
	\centering
	\small
	\begin{minipage}[b]{.38\linewidth}
		\centering
		\includegraphics[width=.9\linewidth]{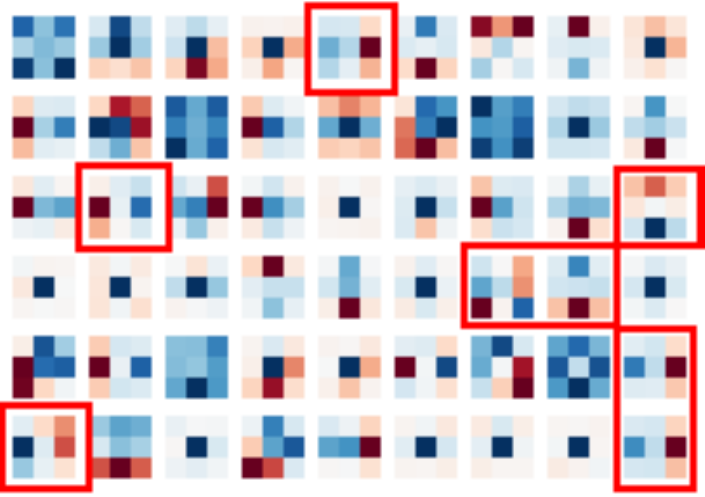}
		\subcaption{\#8/1$\times$3$\times$3}
		\label{fig:sobel_like}
	\end{minipage}
	\begin{minipage}[b]{.075\linewidth}
		\centering
		\includegraphics[width=.5\linewidth]{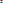}
		\includegraphics[width=.5\linewidth]{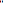}
		\subcaption{}
		\label{fig:sobel_filter}
	\end{minipage}
	\begin{minipage}[b]{.38\linewidth}
		\centering
		\includegraphics[width=.9\linewidth]{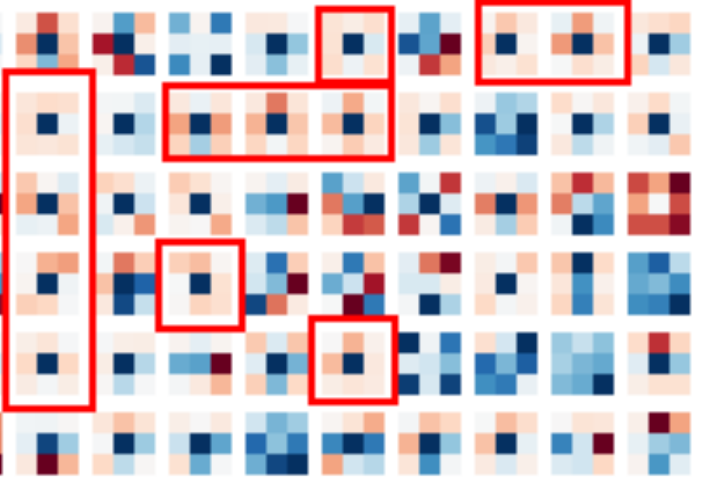}
		\subcaption{\#6/1$\times$3$\times$3}
		\label{fig:laplacian_like}
	\end{minipage}
	\begin{minipage}[b]{.075\linewidth}
		\centering
		\includegraphics[width=.5\linewidth]{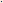}
		\includegraphics[width=.5\linewidth]{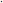}
		\subcaption{}
		\label{fig:laplacian}
	\end{minipage}
	\setlength{\abovecaptionskip}{0.1cm}
	\setlength{\belowcaptionskip}{-0.15cm}
	\captionsetup{font={small}}
	\caption{
		Convolution kernels that are similar to edge detection kernels.
		(a) and (c) are from MobileNetV1. (b) Sobel filter kernels. (d) Laplacian filter kernels.
	}
	\label{fig:filter_kernels_similarly}
\end{figure}

\begin{figure}[t]
	\centering
	\small
	\begin{minipage}{\linewidth}
		\begin{minipage}{.12\linewidth}
			\centering
			\includegraphics[width=\linewidth]{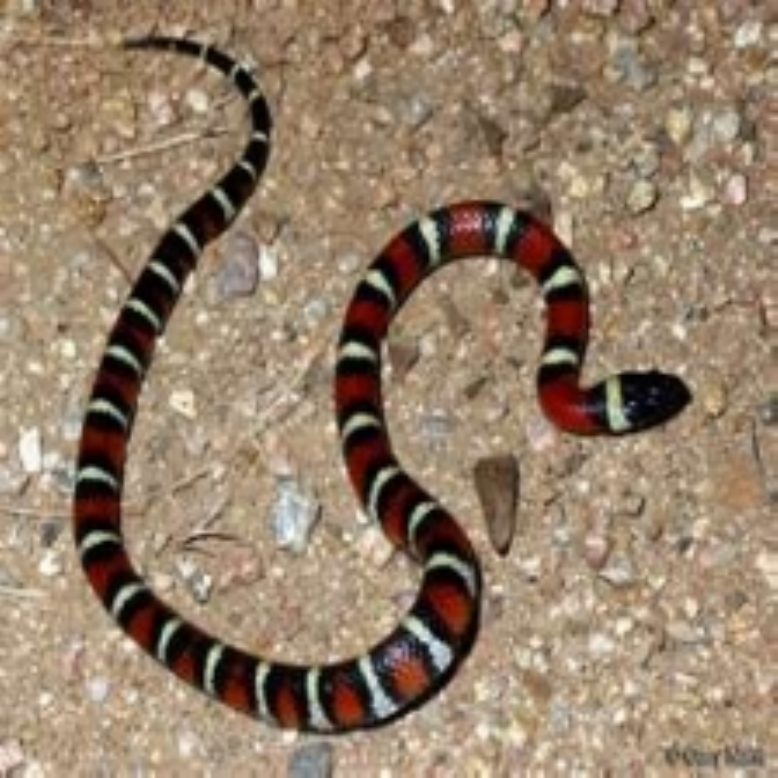}
		\end{minipage}
		\begin{minipage}{.83\linewidth}
			\centering
			\includegraphics[width=\linewidth]{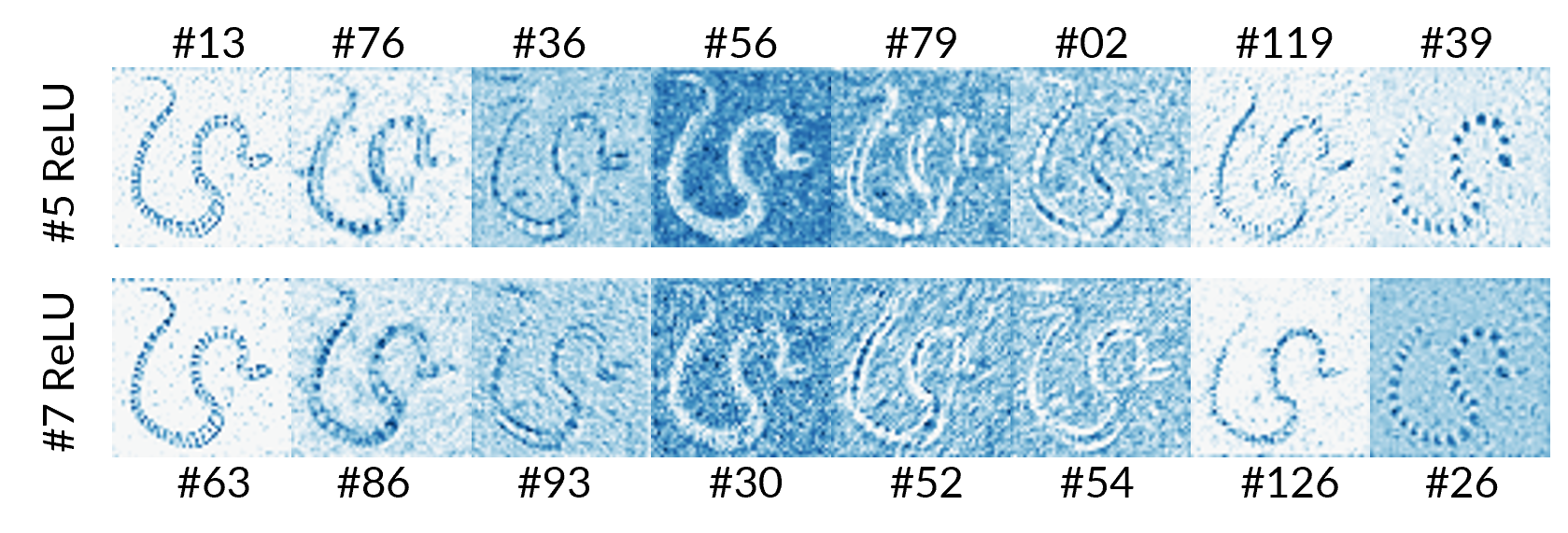}
		\end{minipage}
		\subcaption{Left: the input. Right: feature maps from MobileNetV1.}
	\end{minipage}
	\begin{minipage}{\linewidth}
		\begin{minipage}{.12\linewidth}
			\centering
			\includegraphics[width=\linewidth]{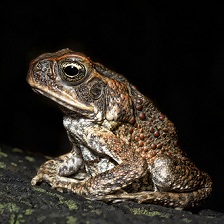}
		\end{minipage}
		\begin{minipage}{.83\linewidth}
			\centering
			\includegraphics[width=\linewidth]{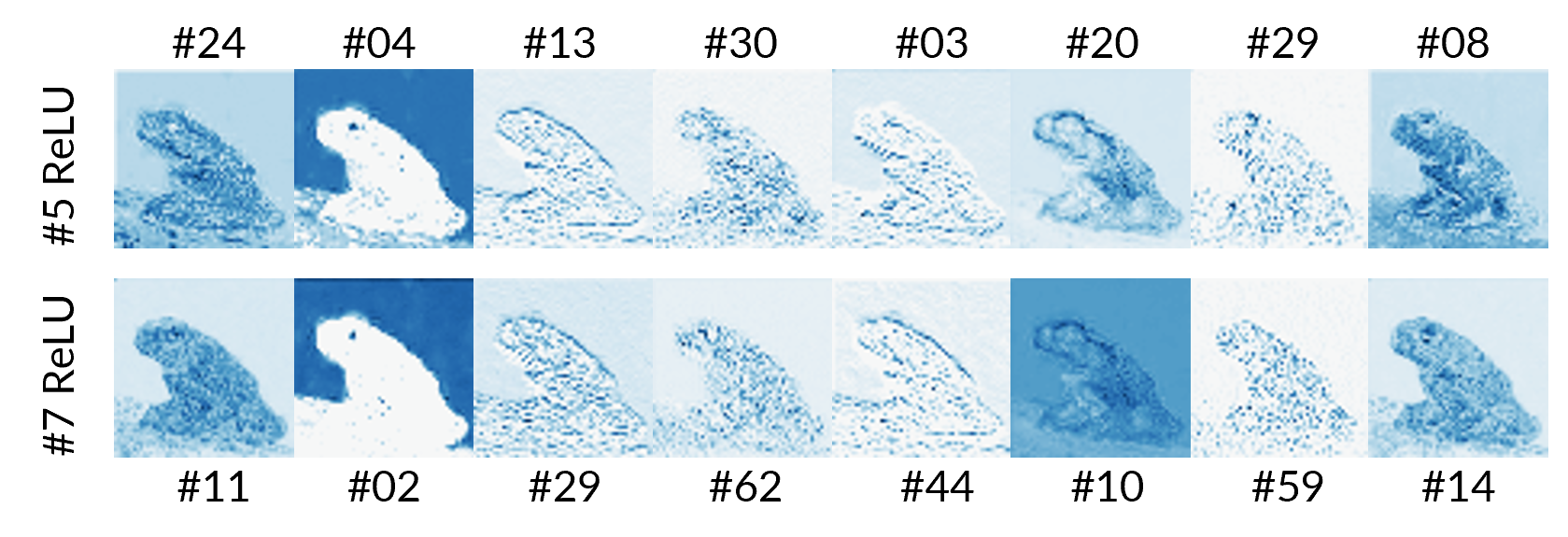}
		\end{minipage}
		\subcaption{Left: the input. Right: feature maps from RegNetX-400MF.}
	\end{minipage}
	\setlength{\abovecaptionskip}{0.1cm}
	\captionsetup{font={small}}
	\caption{
		Similar feature maps from adjacent layers.
		Each row feature maps output from the same layer.
		Here, in \#N, \textit{N} denotes layer number or channel number.
	}
	\label{fig:feature_maps}
\end{figure}

\section{Related Work}

\textbf{DSCs-based Architectures} \textit{Depthwise Separable Convolutions} was introduced in \cite{Sifre2014} and subsequently used in
efficient convolutional neural networks \cite{Howard2017,Sandler2018,Zhang2018,Ma2018,Tan2019a}.
DSCs factorize a standard convolution into a light weight \textit{depthwise convolution} for spatial filtering
and a 1$\times$1 convolution called a heavier \textit{pointwise convolution} for feature generation.
Typical $3 \times 3$ depthwise separable convolutions use between 8 to 9 times less computation than standard convolutions at only a
small reduction in accuracy.

MobileNetV1\cite{Howard2017} employs DSCs to substantially improve computational efficiency.
MobileNetV2\cite{Sandler2018} introduced the \textit{Inverted residual block}, which takes as an input a low-dimensional compressed
representation which is first expanded to high dimension and filtered with a lightweight depthwise convolution.
Features are subsequently projected back to a low-dimensional representation with a linear convolution.
ShuffleNetV1\cite{Zhang2018}, V2\cite{Ma2018} utilizes group convolution and channel shuffle operations to further reduce the complexity.
MobileNetV3\cite{Howard2019}, RegNets\cite{Radosavovic2020} and EfficientNets\cite{Tan2019a} built upon the InvertedResidualBlock structure by introducing lightweight attention modules\cite{Hu2020} based on
squeeze and exctitation into the bottleneck structure.

\textbf{Nyquist-Shannon Sampling Theorem and Shift-Invariant} As reported in \cite{Azulay2019},
the convolutional architecture does not give invariance
since architecture ignores the classical sampling theorem, as small input shifts or translations can cause
drastic changes in the output. Even though blurring before subsampling is sufficient for avoiding aliasing in
linear systems, the presence of nonlinearities may introduce aliasing even in the presence of blur before subsampling.
\cite{Zhang2019} integrates extra classic anti-aliasing to improve shift equivariance of deep networks.

\textbf{Feature Visualization and Attribution} Visualization is a powerful tool to study neural networks.
\textit{Visualization of features} in a fully trained model \cite{Erhan2009,Zeiler2014} makes us see the process of extracting features.
Activation Maximization \cite{Erhan2009,Zeiler2011}, Class Maximization\cite{Yosinski2015} explains what a network is looking for.
Saliency Maps\cite{Simonyan2014}, Guided Backpropagation\cite{Springenberg2015}, Grad-CAM\cite{Selvaraju2020}
explains what part of an example is responsible for the network activating in a particular way.
Network visualization also could give us many intuitive inspirations to design new architectures.

\section{Characteristics of CNNs and Guidelines}
In this section, we study three typical networks constructed by (i) \textit{standard convolutions} such as ResNet-RS,
(ii) \textit{group convolutions} such as RegNet,  (iii) \textit{depthwise separable convolutions} such as MobileNets, ShuffleNetV2 and EfficientNets.
These visualizations demonstrate that M$\times$N$\times$N kernels have distinctly different patterns and distributions at different stages of networks.
What follows are the characteristics of CNNs and guidelines:

\subsection{CNNs can learn to satisfy the sampling theorem}
The previous works\cite{Azulay2019, Zhang2019} always thought that convolutional neural networks ignore the classical sampling theorem,
but we found that \textbf{convolutional neural networks can satisfy the sampling theorem to some extent by learning low-pass filters},
especially the DSCs-based networks such as MobileNetV1 and EfficientNets, as shown in Figure \ref{fig:gaussian}.

\textbf{Standard convolutions/Group convolutions} As shown in Figure \ref{fig:resnet_rs_50} and \ref{fig:regnetx8gf_downsampling},
there are one or more salient N$\times$N kernels like blur kernels in the whole M$\times$N$\times$N kernels,
and this phenomenon also means the parameters of these layers are redundant.
Note that the salient kernels do not necessarily seem like Gaussian kernels.

\textbf{Depthwise separable convolutions} The kernels of strided-DSCs are usually similar to Gaussian kernels, including but not limited to MobileNetV1, MobileNetV2, MobileNetV3, ShuffleNetV2,
ReXNet, EfficientNets. In addition, the distributions of strided-DSC kernels are not Gaussian distributions but Gaussian mixture distributions.

\textbf{Kernels of last convolution layers} Modern CNNs always use global pooling layers before the classifier to reduce the dimension.
Therefore, similar phenomenon is also shown on the last depthwise convolution layers, as shown in Figure \ref{fig:last}.

These visualizations indicate that we should choose depthwise convolutions rather than standard convolutions and group convolutions
in the downsampling layers and last layers.
And further, we could use fixed Gaussian kernels in the downsampling layers.

\subsection{Reuse feature maps between adjacent layers}
\textbf{Identity Kernel and Similar Features Maps} As shown in Figure \ref{fig:identity},
many depthwise convolution kernels only have a large value at the center like \textit{identity kernel} in the middle part of networks.
And convolutions with identity kernels lead to feature maps duplication and computational redundancy since the inputs are just passed to the next layer.
On the other hand, Figure \ref{fig:feature_maps} shows that many feature maps are similar(duplicated) between adjacent layers.

As a result, we could replace partial convolutions with identity mapping.
Otherwise, depthwise convolutions are slow in early layers since they often cannot fully utilize modern accelerators reported in \cite{Ma2018}.
So this method can improve both the parameter efficiency and inference time.

\subsection{Edge detectors as learnable depthwise convolutions}
Edge features contain important information about the images.
As shown in Figure \ref{fig:filter_kernels_similarly}, a large part of kernels approximate to edge detection
kernels, like the Sobel filter kernels and the Laplacian filter kernels. And the proportion of such kernels decreases in
the later layers while the proportion of kernels that like blur kernels increases.

Therefore, maybe the edge detectors could replace the depthwise convolutions in the DSCs-based networks to mix features between different spatial locations.
We will demonstrate that by replacing learnable kernels with edge detection kernels.

\begin{figure}[t]
	\centering
	\begin{minipage}{.32\linewidth}\
		\centering
		\includegraphics[width=.15\linewidth]{figs/fixed/0}
		\includegraphics[width=.15\linewidth]{figs/fixed/1}
		\includegraphics[width=.15\linewidth]{figs/fixed/2}
		\includegraphics[width=.15\linewidth]{figs/fixed/3}
		\subcaption{4 EKs}
	\end{minipage}
	\begin{minipage}{.32\linewidth}
		\centering
		\includegraphics[width=.15\linewidth]{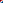}
		\includegraphics[width=.15\linewidth]{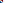}
		\subcaption{Additional 2 Eks}
	\end{minipage}
	\begin{minipage}{.32\linewidth}
		\centering
		\includegraphics[width=.15\linewidth]{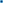}
		\includegraphics[width=.15\linewidth]{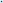}
		\subcaption{Additional 2 GKs}
	\end{minipage}
	\setlength{\abovecaptionskip}{0.1cm}
	\captionsetup{font={small}}
	\caption{
		The 8 unlearnable kernels, including 6 edge detection kernels(EKs) and 2 Gaussian blur kernels(GKs).
		More details can be found in Appendix D.
	}
	\label{fig:filter_kernels}
\end{figure}

\begin{figure}[t]
	\centering
	\small
	\begin{minipage}[b]{.42\linewidth}
		\centering
		\includegraphics[width=.85\linewidth]{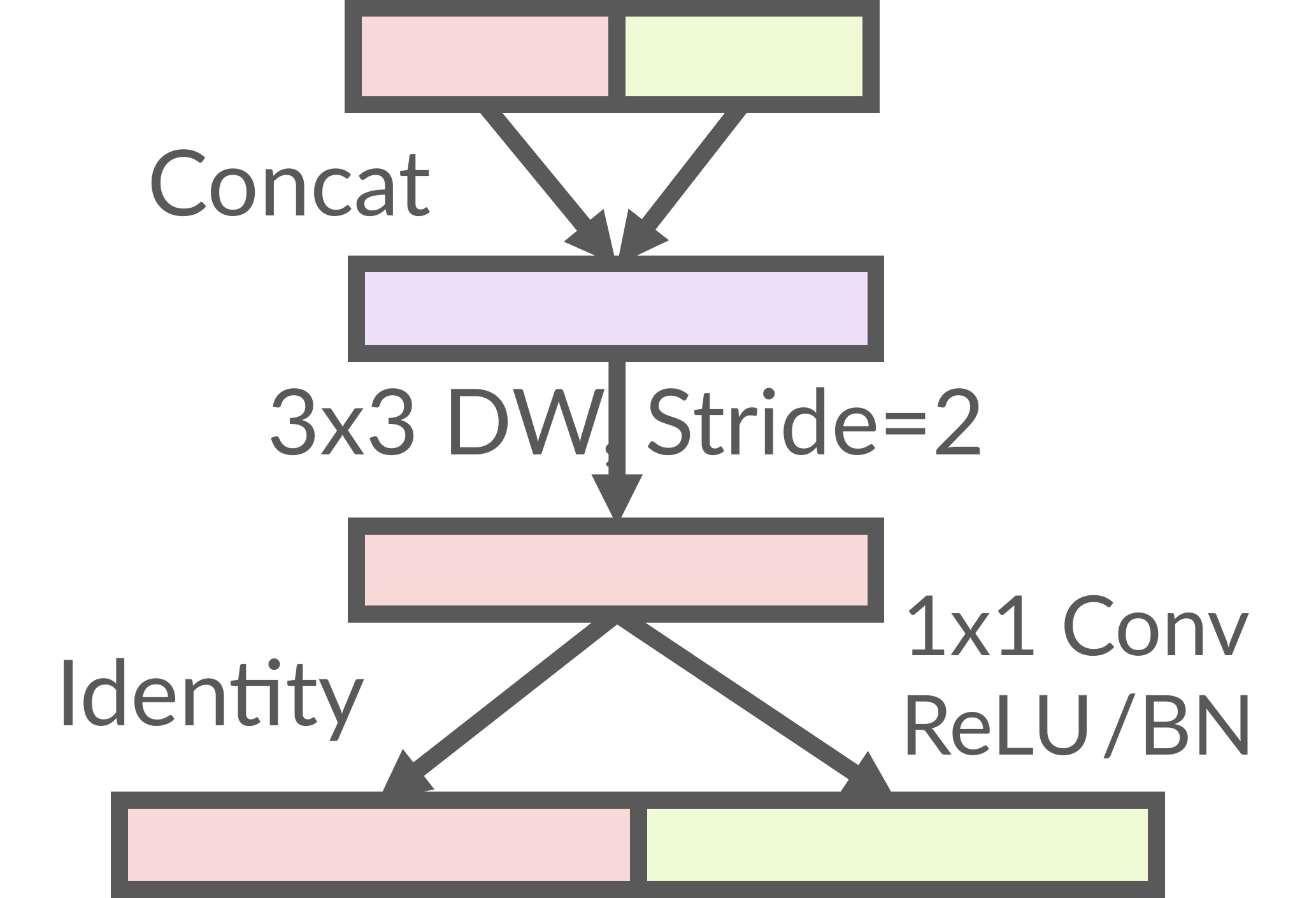}
		\subcaption{DownsamplingBlock}\label{fig:Gaussiandownsamplingblock}
	\end{minipage}
	\begin{minipage}[b]{.42\linewidth}
		\centering
		\includegraphics[width=.85\linewidth]{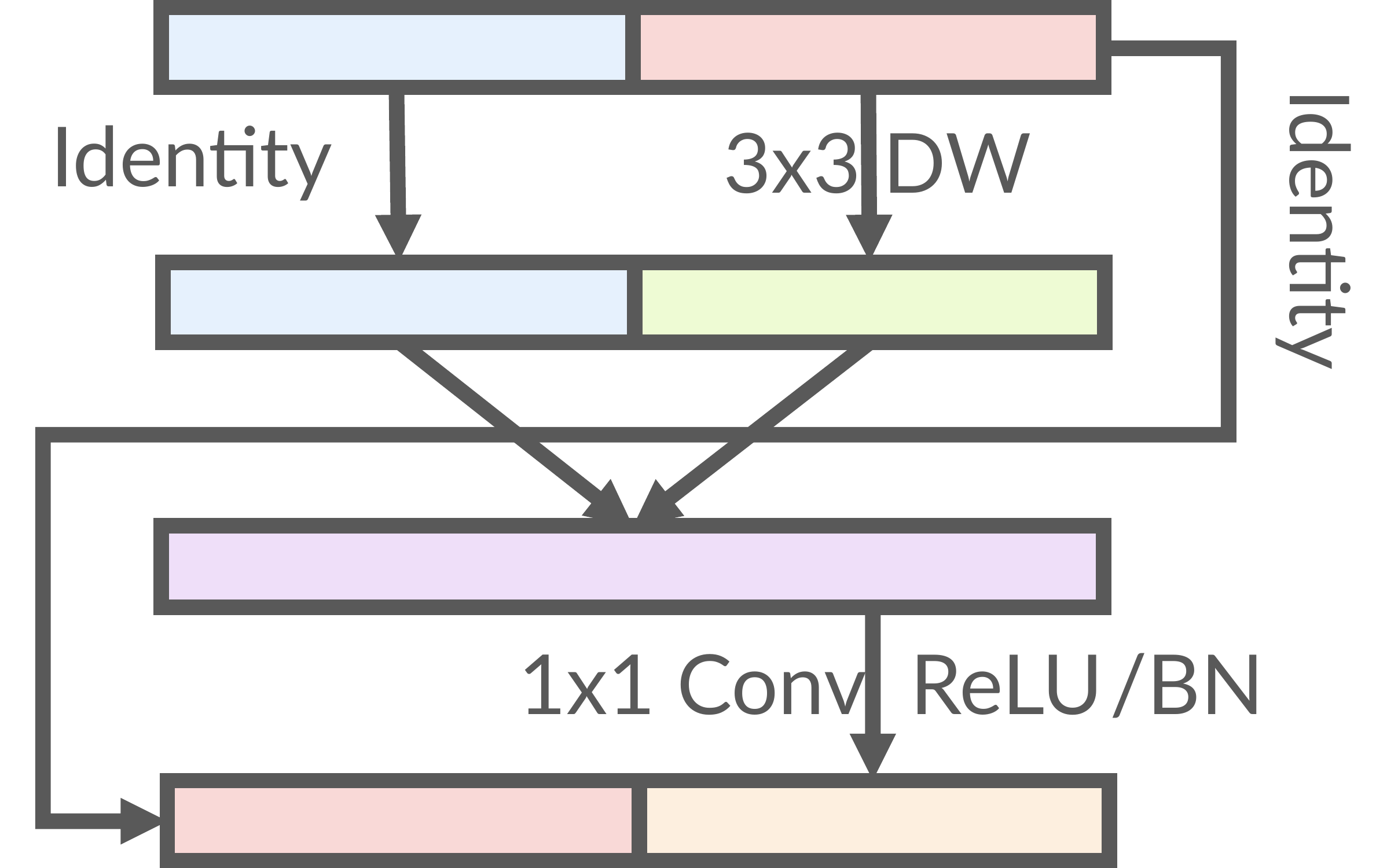}
		\subcaption{HalfIdentityBlock}\label{fig:halfidentityblock}
	\end{minipage}
	\setlength{\abovecaptionskip}{0.1cm}
	\captionsetup{font={small}}
	\caption{
		\textit{DownsamplingBlock} for downsampling and expanding the channels. \textit{HalfIdentityBlock} for increasing the depth.
	}
	\label{fig:arch}
\end{figure}

\section{Network Architecture}
Following these guidelines, we design our parameter-efficient architecture and study the function of depthwise convolutions.
Note that only pointwise convolutions are followed by ReLU and batchnorm for better accuracy and inference speed.

\textbf{DownsamplingBlock} The \textit{DownsamplingBlock} halves the resolution and expands the number of channels.
As shown in Figure \ref{fig:Gaussiandownsamplingblock}, only the expanded channels are generated by the pointwise convolutions
for reusing the features.
The kernels of depthwise convolutions could be randomly initialized or use fixed Gaussian kernels.

\textbf{HalfIdentityBlock}
As shown in Figure \ref{fig:halfidentityblock},
we replace half depthwise convolutions with identity mapping and reduce half pointwise convolutions while \textit{keeping the width of the block}.
\textbf{Note that the right half channels of the input become the left half channels of the output for better features reusing.}

\textbf{VGNet Architecture}
Using the DownsamplingBlock and HalfIdentityBlock, we build our VGNets limited by the number of parameters.
The overall VGNetG-1.0MP architecture is listed in Table \ref{tab:model}.

\begin{table}[t]
	\centering
	\small
	\resizebox{\linewidth}{!}{
		\begin{tabular}{c|c|c|c|c|c}
			\toprule
			Input   & Operator           & Type & Stride & Channels & Layers \\
			\midrule
			$224^2$ & Conv2d             & -    & 2      & 3        & 1      \\
			$112^2$ & DownsamplingBlock  & blur & 2      & 28       & 1      \\
			$56^2 $ & HalfIdentityBlock  & -    & 1      & 56       & 3      \\
			$56^2 $ & DownsamplingBlock  & blur & 2      & 56       & 1      \\
			$28^2 $ & HalfIdentityBlock  & -    & 1      & 112      & 6      \\
			$28^2 $ & DownsamplingBlock  & blur & 2      & 112      & 1      \\
			$14^2 $ & HalfIdentityBlock  & -    & 1      & 224      & 12     \\
			$14^2 $ & DownsamplingBlock  & blur & 2      & 224      & 1      \\
			$7^2 $  & HalfIdentityBlock  & -    & 1      & 368      & 1      \\
			$7^2 $  & SharedDWConv2d,t=8 & -    & 1      & 368      & 1      \\
			$7^2 $  & PointwiseBlock     & -    & 1      & 368      & 1      \\
			$7^2 $  & AvgPool2d          & -    & -      & 368      & 1      \\
			\bottomrule
		\end{tabular}
	}
	\setlength{\abovecaptionskip}{0.1cm}
	\captionsetup{font={small}}
	\caption{\textbf{VGNetG-1.0MP Networks} - \textit{HalfIdentityBlock} and \textit{DownsamplingBlock} are described in Figure \ref{fig:arch}.
		\textit{SharedDWConv2d} just share the same depthwise convolution kernels for $t$ times.}
	\label{tab:model}
\end{table}

\textbf{Variants of VGNet} To further study the impact of the N$\times$N kernels,
several variants of VGNets are introduced: VGNetC, VGNetG, and VGNetF.
\textbf{VGNetC}: All parameters are randomly initialized and learnable.
\textbf{VGNetG}: All parameters are randomly initialized and learnable except the kernels of the DownsamplingBlock.
\textbf{VGNetF}: All parameters of depthwise convolutions are fixed.
The details can be found in Table \ref{tab:variants}.

\begin{table}[t]
	\centering
	\small
	\resizebox{.85\linewidth}{!}{
		\begin{tabular}{c|c|c}
			\toprule
			Variant & \makecell{DownsamplingBlock                       \\ N$\times$N kernels}& \makecell{HalfIdentityBlock \\ N$\times$N kernels} \\
			\midrule
			VGNetC  & random \& learnable         & random \& learnable \\
			VGNetG  & GKs                         & random \& learnable \\
			VGNetF  & GKs                         & EKs \& GKs          \\
			\bottomrule
		\end{tabular}
	}
	\setlength{\abovecaptionskip}{0.1cm}
	\captionsetup{font={small}}
	\caption{
		\textbf{Variants of VGNet}.
		\textit{EKs} denotes edge detection kernels; \textit{GKs} denotes Gaussian kernels.
		EKs and GKs are shown in Figure \ref{fig:filter_kernels}.
	}
	\label{tab:variants}
\end{table}

\section{Experiments}
In this section, we present our experimental setups, the main results on ImageNet
(the results on CIFAR-100 can be found in Appendix B).

\subsection{ImageNet Classification}
The ImageNet ILSVRC2012 dataset contains about 1.28M training images and 50,000 validation images with 1000 classes.
We emphasize that VGNetG models are trained with no regularization except weight decay and label smoothing,
while most networks use various enhancements, such as deep supervision, Cutout, DropPath, AutoAugment, RandAugment, and so on.

\textbf{Training setup} Our ImageNet training settings follow:
SGD optimizer with momentum 0.9;
mini-batch size of 512;
weight decay 1e-4;
initial learning rate 0.2 with 5 warmup epochs;
batch normalization with momentum 0.9;
cosine learning rate decay for 300 epochs;
label smoothing 0.1;
the biases and $\alpha$ and $\beta$ in BN layers are left unregularized.
Finally, all training is done on resolution 224.

\textbf{Results} Table \ref{tab:imagenet} shows the performance comparison on ImageNet,
our VGNetGs achieves better accuracy and parameter efficiency than the MobileNet series and ShuffleNetV2
with more than 30\% parameters reduction and lower inference latency.
In particular, our VGNetG-1.0MP achieves 67.7\% top-1 accuracy with less than 1M parameters and
69.2\% top-1 accuracy with 1.14M parameters.

\begin{table*}[t]
	\small
	\centering
	\begin{tabular}{l|c|cc|cc|cc}
		\toprule
		Model                                         & SE         & \makecell{Top-1                                                                                           \\ Acc.(\%)}    & \makecell{Top-5 \\ Acc.(\%)}    & \makecell{Params \\ (M)}  	& \makecell{Ratio-to\\VGNetG} 	& \makecell{GPU Speed \\ (batches/sec.)} & \makecell{Infer-time \\ (ms)} \\
		\midrule
		MobileNet V2$\times$0.5\cite{Sandler2018}     &            & 63.9            & 85.1          & 1.969          & 1.97$\times$          & 209          & \ 4.77          \\
		MobileNet V3 Small$^\dagger$\cite{Howard2019} & \checkmark & 67.4            & -             & 2.543          & 1.69$\times$          & 196          & \ 5.09          \\
		\textbf{VGNetG-1.0MP(ours)}                   &            & 66.6            & 87.1          & \textbf{0.997} & \textbf{1.00$\times$} & \textbf{228} & \ \textbf{4.37} \\
		\textbf{VGNetG-1.0MP+SiLU(ours)}$^\dagger$    &            & 67.7            & 87.9          & 0.997          & 1.00$\times$          & 226          & \ 4.41          \\
		\textbf{VGNetG-1.0MP+SE(ours)}                & \checkmark & \textbf{69.2}   & \textbf{88.7} & 1.143          & 1.14$\times$          & 107          & \ 9.31          \\
		\midrule
		MobileNet V2$\times$0.75\cite{Sandler2018}    &            & 69.8            & 89.6          & 2.636          & 1.75$\times$          & 194          & \ 5.15          \\
		ShuffleNet V2$\times$1.0\cite{Ma2018}         &            & 69.4            & 88.0          & 2.279          & 1.52$\times$          & 162          & \ 6.15          \\
		\textbf{VGNetG-1.5MP(ours)}                   &            & 69.3            & 88.8          & \textbf{1.502} & \textbf{1.00$\times$} & \textbf{222} & \ \textbf{4.50} \\
		\textbf{VGNetG-1.5MP+SE(ours)}                & \checkmark & \textbf{71.3}   & \textbf{90.1} & 1.702          & 1.13$\times$          & 104          & \ 9.55          \\
		\midrule
		MobileNet V1$\times$1.0\cite{Howard2017}      &            & 70.6            & 88.2          & 4.232          & 2.11$\times$          & 201          & \ 4.97          \\
		MobileNet V2$\times$1.0\cite{Sandler2018}     &            & 72.0            & 91.0          & 3.505          & 1.75$\times$          & 170          & \ 5.87          \\
		MobileNet V3 Large$\times$0.75$^\dagger$      & \checkmark & 73.3            & -             & 3.994          & 1.99$\times$          & 166          & \ 6.01          \\
		\textbf{VGNetG-2.0MP(ours)}                   &            & 71.3            & 90.0          & \textbf{2.006} & \textbf{1.00$\times$} & \textbf{224} & \ \textbf{4.45} \\
		\textbf{VGNetG-2.0MP+SE(ours)}                & \checkmark & \textbf{73.5}   & \textbf{91.4} & 2.345          & 1.17$\times$          & 109          & \ 9.14          \\
		\midrule
		ShuffleNet V2$\times$1.5\cite{Ma2018}         &            & 72.6            & -             & 3.504          & 1.41$\times$          & 157          & \ 6.34          \\
		GhostNet$\times$1.0$^\dagger$\cite{Han2020}   & \checkmark & 73.9            & 91.4          & 5.183          & 2.08$\times$          & 114          & \ 8.75          \\
		RegNetX-400MF\cite{Radosavovic2020}           &            & 72.7            & -             & 5.158          & 2.07$\times$          & 109          & \ 9.12          \\
		RegNetY-400MF\cite{Radosavovic2020}           & \checkmark & 74.1            & -             & 4.344          & 1.74$\times$          & \ 94         & \ 10.62         \\
		\textbf{VGNetG-2.5MP(ours)}                   &            & 72.6            & 90.7          & \textbf{2.493} & \textbf{1.00$\times$} & \textbf{200} & \ \textbf{4.98} \\
		\textbf{VGNetG-2.5MP+SE(ours)}                & \checkmark & \textbf{74.2}   & \textbf{91.8} & 2.922          & 1.17$\times$          & \ \ 97       & 10.31           \\
		\bottomrule
	\end{tabular}
	\setlength{\abovecaptionskip}{0.1cm}
	\captionsetup{font={small}}
	\caption{\textbf{Performance Results on ImageNet}. Our VGNetGs achieve better accuracy with about 30\%$\thicksim$50\% parameters reduction.
		\textit{GPU Speed} and \textit{Infer-time} are measured on RTX6000 GPU with batch size 16.
		\textit{Params} is measured by \textsl{facebookresearch/fvcore} and the \textit{Params} do not include the unlearnable parameters.
		The SEBlocks are used after every pointwise convolutions expect the last one.
		$^\dagger$: used SiLU(also known as the swish function) or HardSwish.
	}
	\label{tab:imagenet}
\end{table*}

\subsection{Ablation study}
In this section, we conduct ablation experiments to gain a better understanding of the impact of the depthwise convolutions.
The ablation experiments are performed on the ImageNet and CIFAR100(see Appendix B).

\textbf{Kernels of downsampling layers} As shown in Table \ref{tab:ablation},
The VGNetG-1.5MP which used the Gaussian blur kernels in the downsampling layers achieves 68.0\% top-1 accuracy, outperforming the
VGNetC-1.5MP by 0.4\% accuracy.

\textbf{Kernels of depthwise convolutions} As shown in Table \ref{tab:ablation},
the VGNetF4, only used 2 Sobel kernels and 2 Laplacian filter kernels instead of all the depthwise convolution kernels,
has about 3\% reduction in accuracy.
The result indicates that CNNs can use edge detectors as learnable depthwise convolutions to mix features.
As mentioned, the last few layers have more kernels like blur kernels.
So the VGNetF2 which used additional Gaussian kernels achieves better accuracy than VGNetF4.

VGNetF1 and VGNetF3 whose last depthwise convolution kernels are learnable only have a 0.8\% reduction in accuracy.

\textbf{Non-linearities and training epochs} Table \ref{tab:ablation_silu} compares the performance of VGNetG-1.5MP according to the number
of training epochs and non-linearities. It can be seen that more epochs and the SiLU non-linearities show better performance.

\begin{table}[ht]
	\small
	\centering
	\resizebox{\linewidth}{!}{
		\begin{tabular}{c|c|c|c|c}
			\toprule
			\makecell{Model                                                             \\ (1.5MP) }& \makecell{Downsampling \\ Kernels } 		& \makecell{N$\times$N \\ Kernels} & \makecell{Last DSC \\ kernels} & \makecell{Top-1 \\ Acc.(\%)}     \\
			\midrule
			VGNetC  & learnable    & learnable            & learnable            & 67.6 \\
			\midrule
			VGNetG  & \textit{GKs} & learnable            & learnable            & 68.0 \\
			\midrule
			VGNetF1 & GKs          & \textit{6 EKs+2 GKs} & learnable            & 66.8 \\
			VGNetF2 & GKs          & 6 EKs+2 GKs          & \textit{6 EKs+2 GKs} & 66.2 \\
			VGNetF3 & GKs          & \textit{4 EKs}       & learnable            & 66.1 \\
			VGNetF4 & GKs          & 4 EKs                & \textit{4 EKs}       & 64.4 \\
			\bottomrule
		\end{tabular}
	}
	\setlength{\abovecaptionskip}{0.1cm}
	\captionsetup{font={small}}
	\caption{\textbf{Abilation study for ImageNet classification}.
		\textbf{EK}: Edge detection Kernel;
		\textbf{GK}: Gaussian Kernel.}
	\label{tab:ablation}
\end{table}

\begin{table}[ht]
	\small
	\centering
	\resizebox{0.65\linewidth}{!}{
		\begin{tabular}{c|c|c|c}
			\toprule
			\makecell{Model }             & SiLU       & Epochs & \makecell{Top-1 \\ Acc.(\%)}     \\
			\midrule
			\multirow{3}{*}{VGNetG-1.5MP} &            & 100    & 68.0            \\
			                              & \checkmark & 100    & 69.1            \\
			                              &            & 300    & 69.3            \\

			\bottomrule
		\end{tabular}
	}
	\setlength{\abovecaptionskip}{0.1cm}
	\captionsetup{font={small}}
	\caption{Impact of non-linearities and training epochs.}
	\label{tab:ablation_silu}
\end{table}

\section{Discussion}
We demonstrated that edge detectors can take the place of learnable depthwise convolution layers in CNNs.
But how these edge features are used is still unclear.

\begin{figure}[t]
	\centering
	\small
	\includegraphics[width=.6\linewidth]{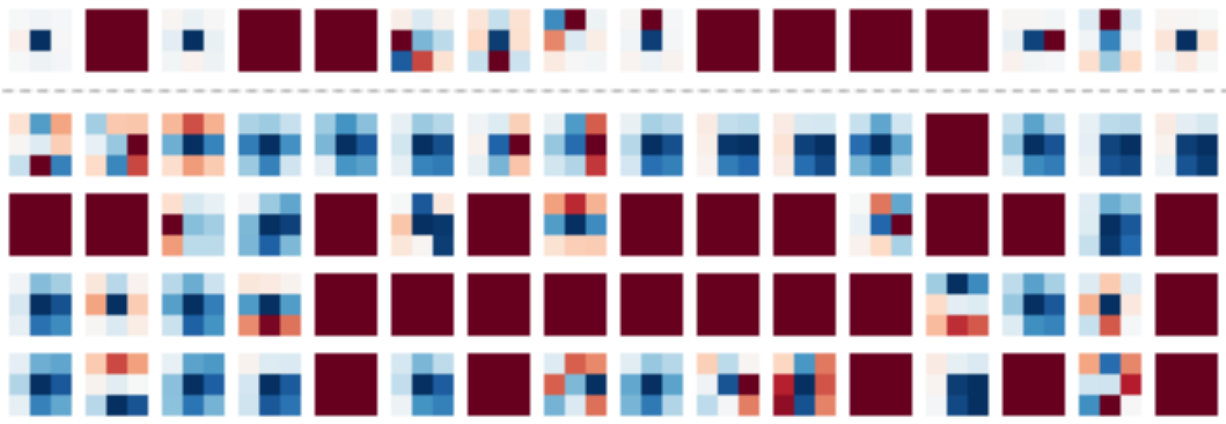}
	\setlength{\abovecaptionskip}{0.1cm}
	\setlength{\belowcaptionskip}{-0.15cm}
	\captionsetup{font={small}}
	\caption{
		MobileNetV3 Large: 3$\times$3 kernels of the second and third depthwise convolution layers.
		As shown in the figure, almost half kernels are zero(red ones).
	}
	\label{fig:zero_kernels}
\end{figure}

Moreover, as shown in Figure \ref{fig:zero_kernels}, MobileNetV3 Small and Large have almost half zero N$\times$N kernels in the \textbf{front} layers.
Maybe we could reduce more parameters and computational complexity in the front layers.

\section{Conclusion}
In this paper, we designed parameter-efficient CNN architecture guided by visualizing the convolution kernels and feature maps.
Based on these visualizations, we proposed VGNets, a new family of smaller and faster neural networks for image recognition.
Our VGNets achieve better accuracy with about 30\%$\thicksim$50\% parameters reduction and lower inference latency than the previous networks.
Finally, we demonstrated that fixed Gaussian kernels and edge detection kernels could replace the learnable depthwise convolution kernels.

\bibliographystyle{IEEEbib}
\bibliography{icme2022}

\end{document}